\documentclass{article}

\usepackage{microtype}
\usepackage{graphicx}
\usepackage{subcaption}
\usepackage{booktabs} 

\usepackage{hyperref}
\usepackage{tabularx}

\usepackage[preprint]{icml2026}

\usepackage{amsmath}
\usepackage{amssymb}
\usepackage{mathtools}
\usepackage{amsthm}
\usepackage{booktabs}
\usepackage{multirow}
\usepackage{colortbl}

\usepackage[capitalize,noabbrev]{cleveref}

\theoremstyle{plain}

\theoremstyle{definition}

\theoremstyle{remark}

\usepackage[textsize=tiny]{todonotes}

\icmltitlerunning{Submission and Formatting Instructions for ICML 2026}

\begin{document}

\twocolumn[
  \icmltitle{Visual Attention Drifts, but Anchors Hold: Mitigating Hallucination in
  \\
   Multimodal Large Language Models via Cross-Layer Visual Anchors}



  \icmlsetsymbol{equal}{*}

  \begin{icmlauthorlist}
    \icmlauthor{Chengxu Yang}{yyy}
    \icmlauthor{Jingling Yuan}{yyy}
    \icmlauthor{Chuang Hu}{comp}
    \icmlauthor{Jiawei Jiang}{comp}
  \end{icmlauthorlist}

  \icmlaffiliation{yyy}{School of Computer Science and Artificial Intelligence, Wuhan University of Technology, China}
  \icmlaffiliation{comp}{School of Computer Science, Wuhan University, China}

  \icmlcorrespondingauthor{Chengxu Yang}{311368@whut.edu.cn}

  \icmlkeywords{Machine Learning, ICML}

  \vskip 0.3in
]



\printAffiliationsAndNotice{}  

\begin{abstract}
  Multimodal Large Language Models often suffer from object hallucination. While existing research utilizes attention enhancement and visual retracing, we find these works lack sufficient interpretability regarding attention drift in final model stages. In this paper, we investigate the layer wise evolution of visual features and discover that hallucination stems from deep layer attention regressing toward initial visual noise from early layers. We observe that output reliability depends on acquiring visual anchors at intermediate layers rather than final layers. Based on these insights, we propose CLVA, which stands for Cross-Layer Visual Anchors, a training free method that reinforces critical mid layer features while suppressing regressive noise. This approach effectively pulls deep layer attention back to correct visual regions by utilizing essential anchors captured from attention dynamics. We evaluate our method across diverse architectures and benchmarks, demonstrating outstanding performance without significant increase in computational time and GPU memory.
\end{abstract}

\section{Introduction}
The rapid development of Multimodal Large Language Models (MLLMs) has revolutionized artificial intelligence by enabling seamless integration of visual and textual information\cite{chen2023shikra,dai2023instructblip}. These models demonstrate impressive performance across various tasks, including Visual Question Answering (VQA), image captioning, and multimodal reasoning\cite{chen2024internvl}. However, MLLMs are often hindered by hallucination issues, where generated outputs deviate from or fabricate details inconsistent with the provided visual content\cite{huang2024opera,park2025convis}. This problem significantly undermines the models' reliability and trustworthiness in practical applications.



Numerous research efforts are dedicated to mitigating hallucinations in MLLMs without retraining; these methods fall into three main categories: (i) contrastive decoding techniques\cite{wan2025only,wang2025ascd} compare outputs from different decoding paths to suppress hallucinations, yet they double the inference cost; (ii) attention modification approaches\cite{yin2025clearsight,chen2025mitigating} enhance the model's focus on visual content by altering attention weights, but they risk amplifying hallucinatory elements; and (iii) layer injection strategies\cite{zou2024look,wang2024mllm} propagate early-layer representations to later layers for better grounding, although their interpretability remains underexplored.

Prior studies have observed that factual tokens in MLLM outputs tend to decrease in later layers while hallucinatory tokens increase\cite{li2025hidden}. Consequently, existing approaches mitigate this issue by weighting later-layer logits based on early-layer logits or employing Visual Retracing to allow models to reuse decoded content from earlier stages. However, the underlying reasons for the decline in factual content accuracy in later layers remain unexplored. Some works suggest that MLLMs possess implicit knowledge of where to attend in images\cite{zhangmllms,li2025token}. In this work, we visualize multi-head attention to visual content through heatmaps and conduct detailed analyses of attention scores. Our findings provide deeper insight into this phenomenon on two fronts: first, visually sensitive heads demonstrate precise and concentrated focus on query-relevant regions in mid-layers; second, this focus diverges in later layers, scattering toward regions remarkably similar to those attended in early layers by visually insensitive heads.


Building on this discovery, we propose the Cross-Layer Visual Anchors (CLVA), a training-free plug-and-play approach designed to enhance visual grounding in MLLMs. During the forward pass, CLVA first distinguishes visually sensitive heads from visually insensitive heads based on their attention to image tokens. It then extracts robust visual anchor patterns: concentrated regions from mid-layer visually sensitive heads and divergent noisy regions from early-layer visually insensitive heads. In later layers, CLVA adaptively refines text-to-image attention by amplifying focus on the concentrated anchor regions while suppressing attention to the divergent ones. This selective refinement effectively counters the observed attention divergence, providing a clear mechanistic explanation for the factual content degradation in deeper layers. Empirical evaluations across multiple hallucination benchmarks confirm the effectiveness of CLVA, yielding consistent improvements in factual accuracy. At the same time, the module introduces minimal computational overhead and preserves near-original inference speed.

In summary, our key contributions are as follows:
\begin{itemize}
       \item We explore the progressive decline in factual grounding within MLLMs. Multi-head attention heatmaps and attention score analyses reveal that attention in later layers collapses toward the distribution patterns of early insensitive heads.
    \item We propose the CLVA. This module employs a visual anchor mechanism to extract robust patterns from early visually insensitive heads and mid-layer visually sensitive heads for targeted refinement in later layers.
    \item We develop CLVA as a training-free solution. The design supports plug-and-play integration, avoids multi-pass decoding, ensures broad compatibility across models, and adds negligible computational overhead.
    \item We perform comprehensive evaluations on hallucination benchmarks. Results confirm gains in factual accuracy and visual grounding.
\end{itemize}


\section{Related Work}
\subsection{Multimodal Large Language Models}
Multimodal Large Language Models (MLLMs) integrate visual perception and language understanding by connecting a pre-trained vision encoder to a large-scale language decoder\cite{devlin2019bert,li2022blip}. Early models such as BLIP-2\cite{dai2023instructblip} and miniGPT-4\cite{zhuminigpt} introduced intermediate modules like the Q-Former\cite{wadekar2024evolution} to bridge modalities effectively. More streamlined approaches, exemplified by LLaVA\cite{liu2024improved}, use simple linear mappings to align visual tokens directly into the language space, reducing complexity while maintaining strong results. Recent models, including LLaVA-Next\cite{liu2024llavanext} and GLM4V\cite{wang2024cogvlm}, adopt a two-stage training process: initial pre-training for modality alignment followed by instruction tuning to refine prompt-following behavior. Despite these innovations, hallucinations persist as a core limitation, where generated text misaligns with visual inputs by fabricating unsupported details.


\subsection{Hallucination Mitigation in MLLMs}

Hallucination in language models originated in unimodal LLMs, where it describes generated content that deviates from factual knowledge or user instructions\cite{huang2025survey,yang2025heaven,bai2024hallucination}. MLLMs inherit this problem and extend it to the visual domain, producing text that misaligns with or fabricates details absent from the input image\cite{jiang2025devils,cho2025you}. Numerous mitigation strategies require additional training. These include curating more robust instruction datasets, applying reinforcement learning with human or AI feedback, and enhancing model architectures\cite{zhu2025popen,yu2025rlaif}. Although such methods yield notable improvements, they demand extensive data preparation and costly fine-tuning.

Training-free approaches provide a resource-efficient alternative by intervening only at inference time. Contrastive decoding methods, such as VCD\cite{VCD}, ICD\cite{ICD}, and IMCCD\cite{IMCCD}, generate parallel output distributions from original and perturbed inputs to favor factual content, though they typically double inference cost due to multiple forward passes. Clearsight\cite{yin2025clearsight} amplifies visual signals in middle-layer fusion to overcome insufficient visual attention. VisFlow\cite{tang2025not} modulates attention dynamics at token and head levels to emphasize salient visual regions while suppressing linguistic bias. MemVR\cite{zou2024look} introduces a look-back mechanism that conditionally re-injects visual tokens as key-value memory upon detecting high uncertainty. DeCo\cite{wang2024mllm} dynamically corrects later-layer logits by incorporating knowledge from preceding layers. These approaches commonly exploit the stronger visual attention observed in mid-layers to counteract degradation in deeper layers. However, they offer limited insight into internal model dynamics and fail to investigate why factual accuracy declines progressively across layers, particularly the role of cross-layer attention divergence. CLVA bridges this gap with an explicit mechanistic explanation and targeted refinement grounded in patterns from visually sensitive and visually insensitive heads.


\section{Motivation}

\subsection{Functional Diversity across Attention Heads} 

Our investigation begins with an empirical analysis of the cross-modal attention mechanisms within the LLaVA1.5-7B\cite{liu2024improved}, which consists of $L=32$ Transformer layers, each containing $H=32$ attention heads. We aim to quantify the degree to which different heads allocate focus toward visual information during the autoregressive decoding process.Following the VisFlow\cite{tang2025not}, for each layer $l \in \{1, \dots, L\}$ and head $h \in \{1, \dots, H\}$, we compute the visual attention intensity $\Phi_{h}^{(l)}$ as the mean attention weight assigned from the set of text tokens $T_{txt}$ to the visual token sequence $T_{vis}$:

\begin{figure}[t]
    \centering
    \includegraphics[width=\linewidth]{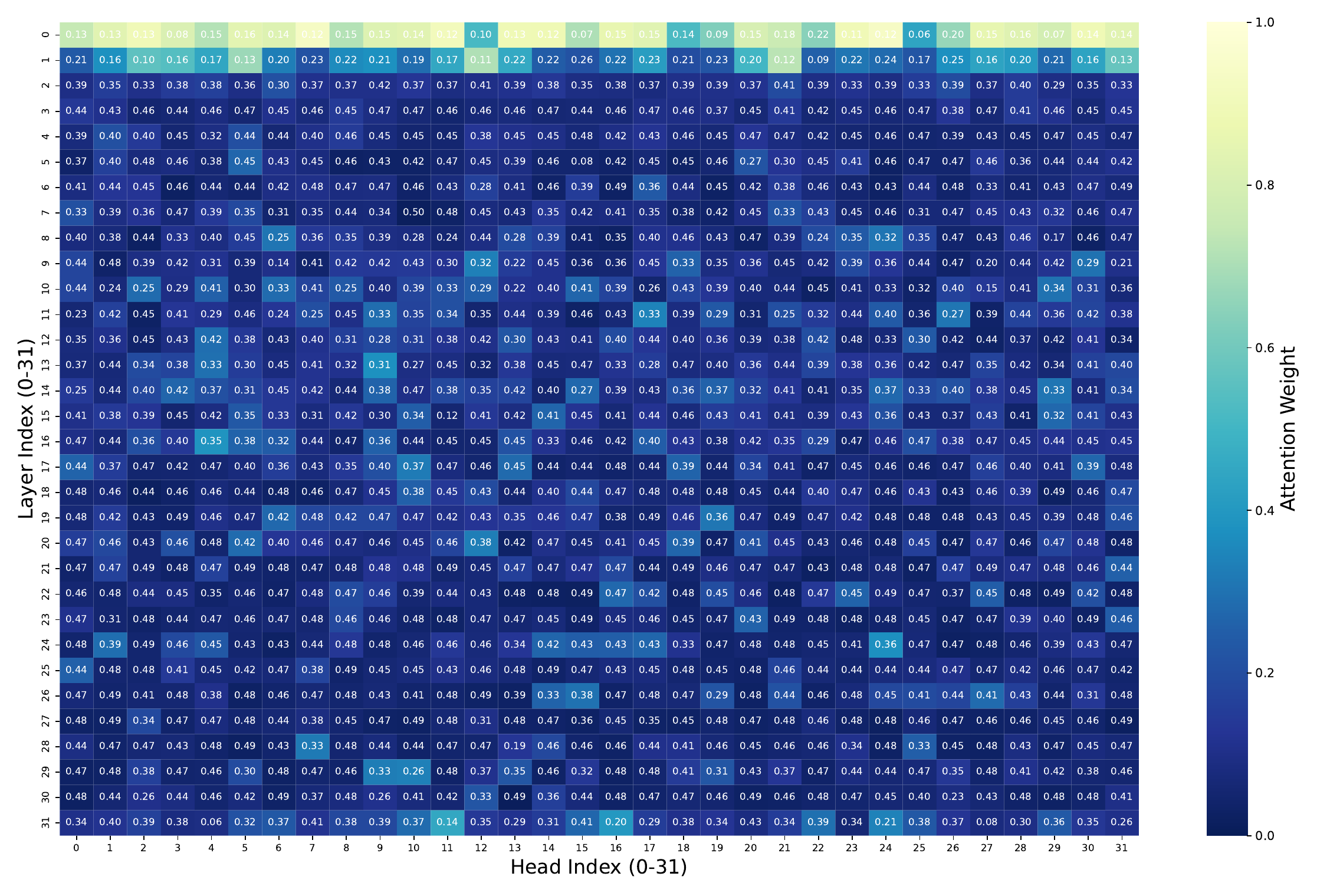}
    \caption{Matrix visualization of visual attention intensity across 32 layers and 32 heads in LLaVA. Colors represent the average attention weights to visual tokens $  T_{vis}  $, while numbers denote the average attention weights to system and text prompt tokens.}
    \label{fig:matrix_vis}
\end{figure}

\begin{equation}
\label{eq:intensity}
\Phi_{h}^{(l)} = \frac{1}{|T_{txt}| } \sum_{i \in T_{txt}} \sum_{j \in T_{vis}} \mathbf{A}_{h}^{(l)}(i, j)
\end{equation}
where $\mathbf{A}_{h}^{(l)}$ denotes the cross-modal attention sub-matrix of the $h$-th head at layer $l$.

As illustrated in Figure \ref{fig:matrix_vis}, we visualize the scores calculated by Eq.\ref{eq:intensity} in a $32 \times 32$ intensity matrix, where each cell $(l, h)$ represents the average grounding strength of a specific head at a given layer. The visualization reveals a distinct sparsity pattern: visual grounding is not a distributed property across the entire architecture but is instead concentrated within a specialized subset of heads, and this pattern varies across layers. Based on this observation, we categorize the heads into visually sensitive heads ($\mathcal{H}_{sens}$) and visually insensitive heads ($\mathcal{H}_{insens}$) through a distribution-based thresholding mechanism:

\begin{equation}
\label{eq:h_sens}
\mathcal{H}_{sens}^{(l)} = \{ h \in H \mid \Phi_{h}^{(l)} > \mu_{\Phi}^{(l)} + \lambda_{vis} \sigma_{\Phi}^{(l)} \}
\end{equation}
\begin{equation}
\label{eq:h_insens}
\mathcal{H}_{insens}^{(l)} = \{ h \in H \mid \Phi_{h}^{(l)} < \mu_{\Phi}^{(l)} - \lambda_{vis} \sigma_{\Phi}^{(l)} \}
\end{equation}

where $\mu_{\Phi}^{(l)}$ and $\sigma_{\Phi}^{(l)}$ denote the mean and standard deviation of visual attention intensity within layer $l$, respectively. The $\lambda_{vis}$ is a hyperparameter that adjusts the selection threshold for functional heads, which we set to $1$ in our experiments. This formalization provides a principled basis for isolating functional heads that drive spatial grounding from those that capture modality-agnostic noise.

\begin{figure}[t]
\centering
\includegraphics[width=\linewidth]{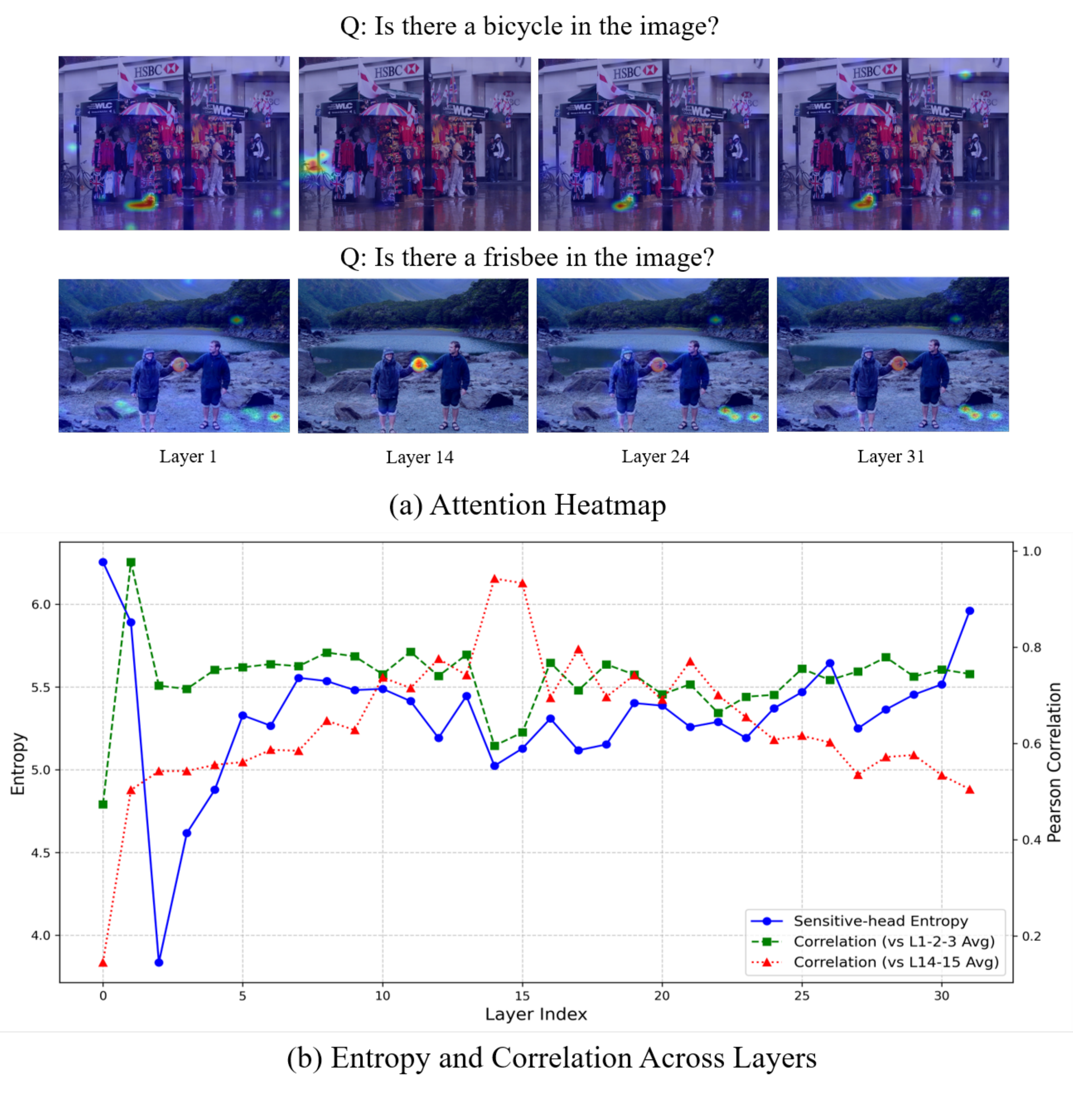} 
\caption{Attention analysis on LLaVA-1.5-7B. (a) Heatmap visualization of attention patterns across different layers. (b) Statistical metrics where the blue line represents the average attention entropy of vision sensitive heads. The green and red lines denote the Pearson correlation coefficients between the attention distributions of each layer and the mean attention of insensitive heads in initial layers and sensitive heads in intermediate layers, respectively.}
\label{fig:llava_analysis}
\end{figure}


\subsection{Visual Anchors and Noise Recurrence}

\textbf{Dual-Anchor Identification via Visualization.} Drawing inspiration from recent studies on the internal mechanisms of MLLMs \cite{zhang2023visual,zhang2024exploring,wu2024v}, which suggest that models inherently possess the capability to identify task-relevant regions despite eventually generating hallucinations, we investigate the evolution of this ``internal knowledge'' across the layers of the model. By conducting a qualitative evaluation on a diverse set of samples from the COCO dataset\cite{lin2014microsoft}, we identify a dual-anchor mechanism governing the model's focus during the forward pass. We derive the visual saliency map $\mathbf{M}^{(l, \mathcal{H})}$ for a layer $l$ and a head subset $\mathcal{H}$ by averaging the attention weights from the last query token to the visual token sequence $T_{vis}$:

\begin{equation}
\label{eq:saliency_extraction}
\mathbf{M}^{(l, \mathcal{H})} = \left[ \frac{1}{|\mathcal{H}|} \sum_{h \in \mathcal{H}} \mathbf{A}_{h}^{(l)}(-1,j) \right]_{j \in T_{vis}}
\end{equation}

We generate heatmaps for 100 images from the MSCOCO dataset to visualize the attention patterns. As shown in Figure \ref{fig:llava_analysis}(a), we observe that at intermediate depths, the visually sensitive heads consistently produce \textbf{Positive Visual Anchors} that accurately pinpoint task-relevant semantic regions, demonstrating that the model effectively ``knows where to look''. Conversely, the initial layers establish \textbf{Negative Visual Anchors} characterized by uninformative spatial biases or background noise. While the positive anchors provide grounded visual evidence, this sharp focus is not maintained; as the computation moves toward the final layers, the attention undergoes a structural shift, abandoning these positive anchors and reverting to the negative anchors inherent in the model's initial layers.


\textbf{Information Decay and Systematic Noise Recurrence.} To provide a formal basis for the existence and subsequent decay of these visual anchors, we analyze the attention distribution across each layer of the model through multiple metrics and calculate the mean values over the aforementioned 100 images. Specifically, we utilize \textbf{Attention Entropy} $H$ to track the stability and dispersion of the visual anchors, calculated as:
\begin{equation}
\label{eq:entropy}
H(\mathbf{M}^{(l, \mathcal{H})}) = -\sum_{j \in T_{vis}} \mathbf{M}_{j}^{(l, \mathcal{H})} \log(\mathbf{M}_{j}^{(l, \mathcal{H})})
\end{equation}
Furthermore, to verify the nature of the deep-layer drift, we calculate the \textbf{Pearson correlation coefficient $r$} between the attention maps of the all layers and the Vision distributions (both the positive and negative anchors):
\begin{equation}
\label{eq:pearson}
r(l) = \frac{\text{Cov}(\mathbf{M}^{(l, \mathcal{H}_{all})}, \mathbf{M}^{(1, \mathcal{H}_{insens})})}{\sigma_{\mathbf{M}^{(l, \mathcal{H}_{all})}} \sigma_{\mathbf{M}^{(1, \mathcal{H}_{insens})}}}
\end{equation}

The statistical metrics in Figure \ref{fig:llava_analysis}(b) reveal a systematic \textbf{Deep-Layer Attention Drift} characterized by the model's shifting internal focus. While the Attention Entropy exhibits fluctuations across all depths, the intermediate layers function as the peak grounding center where the model establishes Positive Visual Anchors, marked by a combination of low entropy and high semantic alignment. However, this grounded focus significantly diverges as the computation progresses. This drift is evidenced by the divergent trends in Pearson correlation: the correlation with intermediate positive anchors steadily declines in the final stages, while the correlation with early-layer noise priors simultaneously rises. Notably, even in instances where deep layers exhibit lower entropy than the mid-layers, these values often coincide with a high correlation to Layer 1 patterns. This suggests that the model's late-stage focus is not regaining visual evidence, but is instead being steered toward incorrect regions dictated by linguistic biases. These observations prove that the model’s internal visual evidence is overwritten by a systematic regression toward initial-layer spatial noise. This phenomenon directly motivates our CLVA method, which intervenes in the drifted deep-layer regime by reinforcing positive anchors and suppressing the recurrence of negative anchors.


\section{Methodology}
\begin{figure*}[t]
\centering
\includegraphics[width=\linewidth]{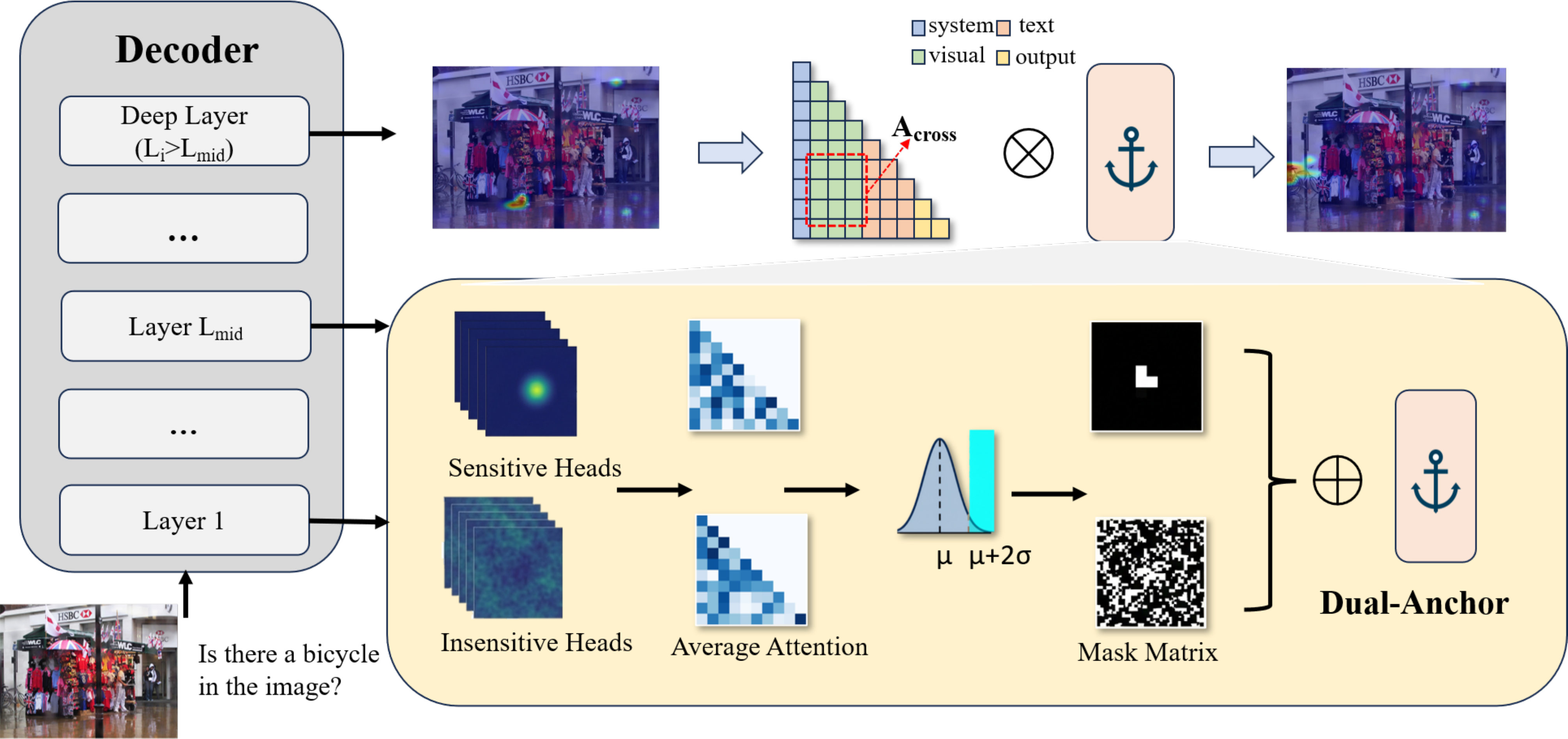} 
\caption{Architecture of CLVA. Based on our attention analysis, we propose a two-step process to anchor drifting attention, where positive anchoring anchors core visual semantic features while negative anchoring excludes background noise using insensitive heads.}
\label{fig:CLVA}
\end{figure*}

Based on the above findings, we further explore the hallucination mechanism in MLLMs from the perspective of interpretability. Our exploration reveals that the model effectively captures task-relevant regions during intermediate stages, but the subsequent drift in deeper layers represents a systematic regression to early-layer noise priors rather than random dispersion. This suggests that as reasoning depth increases, the model tends to overwrite internal visual evidence with intrinsic spatial biases. To address this, we propose Cross-Layer Visual Anchors (CLVA), a training-free intervention that reinforces grounding knowledge from intermediate layers while actively stripping away recurring noise from the initial layers to restore visual faithfulness. The overview of our method is illustrated in Figure \ref{fig:CLVA}.

\subsection{Preliminaries}
MLLMs typically comprise a visual encoder, a cross-modal interface, and a language decoder\cite{zhang2024debiasing,leng2024mitigating}. During autoregressive inference, the input image is encoded into $n$ visual tokens $T_{vis} = \{v_i\}_{i=1}^n$. These visual tokens are then concentrated to generate the input tokens $[T_{sys}, T_{vis}, T_{txt}]$ for the language decoder, where $T_{sys} = \{t_i\}_{i=0}^{m_b}$ represents the system prompt tokens and $T_{txt} = \{t_i\}_{i=m_b+1}^m$ denotes the instruction tokens. In each self-attention layer of the language decoder, the input hidden states are projected into queries $Q$, keys $K$ and values $V$ of dimension $(n+m) \times d$. The attention matrix $\mathbf{A} \in \mathbb{R}^{(n+m) \times (n+m)}$ is estimated by:
\begin{equation}
\mathbf{A}_{logits} = \frac{QK^\top}{\sqrt{d}}, \quad \mathbf{A} = \text{softmax}(\mathbf{A}_{logits})
\end{equation}
where $\mathbf{A}_{logits}$ is the attention logits before softmax. The matrix $\mathbf{A}$ represents the relevance scores between all token pairs. In the CLVA framework, we specifically intervene in the attention weights directed from the $T_{txt}$ to the $T_{vis}$. By applying contrastive anchoring to this attention, we steer the model's focus back to the relevant visual evidence identified in the intermediate layers while suppressing the influence of early-layer spatial priors.

\subsection{Dual-Anchor Target and Saliency Masking}

To operationalize the attention intervention, we define two contrastive anchors derived from specialized head behaviors. The positive anchor $\mathbf{M}_{pos}$ aggregates attention from vision sensitive heads $\mathcal{H}_{sens}$ at an intermediate layer $l_{mid}$ to capture grounded semantic distributions, while the negative anchor $\mathbf{M}_{neg}$ is extracted from vision-insensitive heads $\mathcal{H}_{insens}$ at the initial layer to characterize inherent spatial biases. For each anchor $k \in \{pos, neg\}$, we employ a statistical outlier detection mechanism to isolate significant regions. Specifically, for each visual token $j \in T_{vis}$, the Z-score $z_j^{(k)}$ is computed as:

\begin{equation}z_j^{(k)} = \frac{\mathbf{M}_{k}(j) - \mu_{k}}{\sigma_{k} + \epsilon}, \quad j \in T_{vis}
\end{equation}

where $\mu_{k}$ and $\sigma_{k}$ denote the mean and standard deviation of the saliency map $\mathbf{M}_{k}$ across all visual tokens, and $\epsilon$ is a small constant for numerical stability. The resulting binary masks $\mathbf{Z}_{pos}$ and $\mathbf{Z}_{neg}$ are then formulated as the support of spatial outliers exceeding a significance threshold $\tau$:
\begin{equation}\mathbf{Z}_{k}(j) = \mathbb{I}(z_j^{(k)} > \tau),\quad j \in T_{vis}
\end{equation}
In this formulation, $\mathbb{I}$ is the indicator function. These dual masks provide categorical geometric constraints to guide the attention re-anchoring process in the deeper layers.

\subsection{Attention Re-anchoring}
The intervention process operationalizes the dual-anchor strategy by performing layer-wise rectification within the drift region. Specifically, at each decoding step within the affected layers, the attention matrix $\mathbf{A}$ is adjusted to enhance the model's focus on relevant visual regions while actively suppressing the erroneous drift toward early-layer noise. For each textual token $i \in T_{txt}$ and visual token $j \in T_{vis}$, the attention weights are modulated by the dual anchors as follows:
\begin{equation}\tilde{\mathbf{A}}(i, j) = \mathbf{A}(i, j) \cdot \left(1+\alpha \mathbf{Z}_{pos}(j) - \beta \mathbf{Z}_{neg}(j) \right)
\end{equation}
where $\alpha$ and $\beta$ are hyperparameters that govern the intensity of positive reinforcement and negative suppression, respectively. To preserve the probabilistic interpretation of the attention mechanism, the modified weights undergo a secondary normalization:
\begin{equation}\hat{\mathbf{A}}(i, j) = \frac{\tilde{\mathbf{A}}(i, j)}{\sum\limits_{j} \tilde{\mathbf{A}}(i, j)}, \quad i \in T_{txt}
\end{equation}
 
This dual-stage strategy enables the model to dynamically recalibrate its cross-modal alignment without additional training, ensuring that the language decoder remains anchored to high-fidelity visual evidence and effectively mitigating the emergence of hallucinated content in complex reasoning tasks.

\subsection{Theoretical Analysis}


The re-anchoring of attention weights directly reshapes the output representation $O$, which serves as the core signal for autoregressive decoding. In the Transformer architecture, the output is computed as $ O = AV $, where $ A $ is the attention matrix and $ V $ the value matrix. By partitioning the sequence into linguistic and visual tokens, this can be decomposed as:
\begin{equation}O_i = \underbrace{\sum_{j \in T_{sys} \cup T_{txt}} \hat{\mathbf{A}}(i, j) V_j}_{O_{P}} + \underbrace{\sum_{k \in T_{vis}} \hat{\mathbf{A}}(i, k) V_k}_{O_{V}}
\end{equation}
where $  O_i  $ is the output for query token $  i  $. In shallow layers, the visual component $  O_V  $ typically dominates, reflecting initial modality fusion. In deeper layers, however, LVLMs naturally exhibit perceptual collapse where $  A(i, k)  $ for visual tokens $  T_{vis}  $ diminishes toward zero. Consequently, the linguistic component $  O_P  $ dominates the output, forcing the hidden state to be updated primarily by language priors stored in the system prompt $  T_{sys}  $ and preceding context $  T_{txt}  $. CLVA intervenes by replacing the original weights with the re-anchored weights $  \hat{A}  $. This intervention fundamentally alters the composition of the hidden state through two direct effects:
\begin{itemize}
    \item \textbf{Suppression of Language Priors}: As the mass of $\hat{\mathbf{A}}$ for visual tokens increases, the weights for linguistic tokens $T_{sys} \cup T_{txt}$ are proportionally reduced due to the re-normalization. This dampens the magnitude of $O_P$, thereby mitigating hallucinations driven by excessive reliance on language priors.
    \item \textbf{Visual Signal Purification}: By concentrating the attention mass on regions validated by $\mathbf{Z}_{pos}$ and pruning those in $\mathbf{Z}_{neg}$, the visual component $O_{vis}$ is refreshed with high-fidelity semantic evidence.
\end{itemize}

The intervention framework is compatible with various LVLM paradigms without structural modification. For linear projection, the re-anchoring is applied to the self-attention layers of the decoder. In models utilizing a Q-Former or visual abstractor, the strategy is implemented within the cross-attention mechanism bridging the visual encoder and the LLM. 

\begin{table*}[t!]
\centering
\caption{Results on the POPE benchmark. Results are averaged across the MS-COCO, A-OKVQA, and GQA datasets. The best and second best results are bolded and underlined, respectively. “-” indicates the result is not supported by the released implementation.}
\label{tab:pope}
\resizebox{\textwidth}{!}{
\begin{tabular}{llcccccccc}
\toprule
\multirow{2}{*}{Evaluation} & \multirow{2}{*}{Methods} & \multicolumn{2}{c}{Random} & \multicolumn{2}{c}{Popular} & \multicolumn{2}{c}{Adversarial} & \multicolumn{2}{c}{Average} \\
\cmidrule(lr){3-4} \cmidrule(lr){5-6} \cmidrule(lr){7-8} \cmidrule(lr){9-10} 
 &  & F1-score $\uparrow$ & Accuracy $\uparrow$ & F1-score $\uparrow$ & Accuracy $\uparrow$ & F1-score $\uparrow$ & Accuracy $\uparrow$ & F1-score $\uparrow$ & Accuracy $\uparrow$  \\
\midrule
\multirow{5}{*}{LLaVA-1.5} & Greedy & 87.18 & 88.26 & 84.48 & 85.02 & 81.24 & 81.11 & 84.30 &  84.79 \\
 & Beam Search & 87.26 & 88.35 & 84.50 & 85.03 & 81.21 & 81.09 & 84.32 & 84.82   \\
 & MemVR & 87.83 & 88.40 & 83.20 & 84.37 & 80.61 & 80.53 & 83.88 & 84.43   \\
 & ClearSight & 88.16 & 88.77 & 85.13 & 85.02 & 81.44 & 80.39 & 84.91 & 84.72  \\
 & IMCCD & \underline{88.22} & \underline{88.82} & \underline{85.45} & \underline{86.00} & \underline{82.07} & \underline{81.66} & \underline{85.24} & \underline{85.49}  \\
 \rowcolor{gray!20}
 & CLVA (ours) & \textbf{88.70} & \textbf{89.37} & \textbf{85.82} & \textbf{86.39} & \textbf{82.50} & \textbf{82.19} & \textbf{85.67} &  \textbf{85.98}  \\
\midrule
\multirow{5}{*}{InstructBLIP} & Greedy & 81.25 & 80.75 & 78.17 & 76.60 & 76.12 &  73.71 & 78.51  & 77.02   \\
 & Beam Search & 81.30 & 80.80 & 78.17 & 76.60 & 76.09 & 73.69 & 78.52 &  77.03  \\
 & MemVR & - & - & - & - & - & - & - & -  \\
 & ClearSight  & - & - & - & - & - & - & - & -  \\
 & IMCCD & \underline{85.85} & \underline{86.30} & \underline{81.79} & \underline{81.22} & \underline{78.90} & \underline{77.29} & \underline{82.18} & \underline{81.60}  \\
 \rowcolor{gray!20}
 & CLVA (ours) & \textbf{87.58} & \textbf{87.88} & \textbf{82.53} & \textbf{81.79}& \textbf{79.47} & \textbf{77.72} & \textbf{83.19} & \textbf{82.46}   \\
\midrule
\multirow{5}{*}{Qwen-VL} & Greedy  & 84.88 & 86.34 & 83.45 & 84.68 & 81.93 & 82.86 & 83.42 & 84.62   \\
 & Beam Search  & 84.88 & 86.24 & 83.40 & 85.62 & 81.97 & 82.89 & 83.41 & 84.91  \\
 & MemVR  & 85.75 & 87.30 & \underline{85.19} & 84.73 & 82.07 & 82.90 & 84.33 & 84.97  \\
 & ClearSight  & 85.94 & 87.63 & 84.83 & 85.78 & 81.95 & \underline{82.94} & 84.24 & 85.45  \\
 & IMCCD & \textbf{86.66} & \textbf{87.79} & \textbf{85.50} & \underline{86.05} & \textbf{82.42} & 82.84 & \textbf{84.86} & \textbf{86.56}  \\
 \rowcolor{gray!20}
 & CLVA (ours) & \underline{86.51} & \underline{87.75} & 84.74 & \textbf{86.13} & \underline{82.21} & \textbf{83.45} & \underline{84.48} &  \underline{85.77}  \\
\bottomrule
\end{tabular}
}
\end{table*}

\section{Experiments}
\subsection{Models and Experimental Setup}
\textbf{Model Selection.} We evaluate CLVA on three representative models: LLaVA-1.5\cite{liu2024improved}, Qwen-VL\cite{bai2023qwen}, and InstructBLIP\cite{dai2023instructblip}. We adopt the 7 billion parameters (7B) version for the language model backbone across all three architectures to ensure a consistent comparison. 

\textbf{Baselines.} We compare CLVA against three state-of-the-art training-free methods representing different technical approaches. \textbf{IMCCD}\cite{IMCCD} adopts an improved multimodal contrastive decoding framework to refine output distributions. \textbf{MemVR}\cite{zou2024look} utilizes a visual retracing mechanism to verify generated content against the original visual features. \textbf{ClearSight}\cite{yin2025clearsight} focuses on mitigating hallucinations by amplifying the model's attention toward image tokens during the decoding process.

\textbf{Decoding and Hyper-parameters.} Greedy search is the default decoding protocol unless otherwise specified. For evaluations involving beam search, the beam size is fixed at 5. CLVA utilizes the first layer as the negative anchor, while the positive anchor is set to layer $L/2 - 2$, where $L$ denotes the total number of model layers. These specific anchor layers were selected based on preliminary experiments on the MSCOCO validation set to ensure balanced performance. The amplification factor $\alpha$, the penalty factor $\beta$, and the significance threshold $\tau$ were set to 14, 0.9, and 2, respectively.


\subsection{Hallucination Benchmark Results}
CLVA is evaluated on two primary hallucination benchmarks, POPE\cite{pope} and CHAIR\cite{chair}. These datasets assess the model performance from discriminative probing and generative captioning perspectives, respectively.


\textbf{POPE.} This benchmark frames object hallucination as a binary classification task where models must determine the presence of specific objects. The evaluation involves nine distinct subsets constructed from three source datasets: MSCOCO, AOKVQA, and GQA. Each source is paired with three sampling strategies labeled as Random, Popular, and Adversarial. Each subset contains 500 images with 6 questions per image. Following established protocols, a constraint requiring a one-word answer is appended to each query to ensure consistency. Accuracy and F1 score serve as the primary performance metrics.

\begin{table*}[t!]
\centering
\caption{CHAIR hallucination evaluation results. The best and second best results are bolded and underlined, respectively.}
\label{tab:chair}
\resizebox{\textwidth}{!}{
\begin{tabular}{cccccccccc}
\toprule
\multirow{2}{*}{Methods} & \multicolumn{3}{c}{LLaVA-1.5} & \multicolumn{3}{c}{InstructBLIP} & \multicolumn{3}{c}{Qwen-VL} \\
\cmidrule(lr){2-4} \cmidrule(lr){5-7} \cmidrule(lr){8-10}
 & CHAIR$_i$ $\downarrow$ & CHAIR$_s$ $\downarrow$ & Recall $\uparrow$ & CHAIR$_i$ $\downarrow$ & CHAIR$_s$ $\downarrow$ & Recall $\uparrow$ & CHAIR$_i$ $\downarrow$ & CHAIR$_s$ $\downarrow$ & Recall $\uparrow$ \\
\midrule
Greedy & 53.0 & 15.2&76.7 &57.0 &16.7 & 70.4& 6.6& 7.0& 45.5\\
Beam Search & 55.6&16.0 &76.7 &56.2 &\underline{14.7} &71.4 &6.4 &5.7 &\underline{45.7} \\
MemVR &\underline{49.2} &\underline{12.9} &78.4 &- &- &- &4.7 & 4.0& 45.3\\
ClearSight & 50.0& 13.6& \underline{78.8}& -&- &- & \underline{4.2}& \underline{3.8}& 44.0\\
IMCCD & 52.2& 13.5& 76.6& \underline{55.3}& 17.0& \underline{73.0}&6.6 &5.3 & 41.7\\
\rowcolor{gray!20}
CLVA (ours) & \textbf{48.0}& \textbf{12.7}& \textbf{80.2}& \textbf{50.2}& \textbf{14.0}& \textbf{74.0}& \textbf{3.8}& \textbf{2.7}& \textbf{46.3}\\
\bottomrule
\end{tabular}
}
\end{table*}

\begin{table}[t]
\caption{Results on the hallucination subset of MME. The best performances within each setting are bolded.}
\label{tab:mme}
\resizebox{\columnwidth}{!}{
\begin{tabular}{llccccc}
\toprule
\multirow{2}{*}{\textbf{Model}} & \multirow{2}{*}{\textbf{Decoding}} & \multicolumn{2}{c}{\textbf{Object-level}} & \multicolumn{2}{c}{\textbf{Attribute-level}} & \multirow{2}{*}{\textbf{Total} $\uparrow$} \\
\cmidrule(lr){3-4} \cmidrule(lr){5-6}
 &  & Existence $\uparrow$ & Count $\uparrow$ & Position $\uparrow$ & Color $\uparrow$ &  \\
\midrule
\multirow{6}{*}{LLaVA-1.5} & Greedy & 170.00 & 130.00 &  93.33 & 121.66 & 514.99 \\
 & Beam Search & 170.00 &130.00&93.33& 121.66 & 514.99 \\
 & MemVR & 190.00 & 155.00 & \textbf{133.33} &170.00& 648.33 \\
 & ClearSight & 190.00 & 160.00 & \textbf{133.33} & 160.00 & 643.33 \\
 & IMCCD &\textbf{195.00}  & 138.33 & 123.33 & 155.00 & 611.66 \\
 & CLVA (ours) & \textbf{195.00} & \textbf{163.33} & \textbf{133.33} & \textbf{175.00} & \textbf{666.66} \\
\midrule
\multirow{4}{*}{InstructBLIP} & Greedy & \textbf{185.00} & 60.00 & 50.00 & 120.00 & 415.00 \\
 & Beam Search & 175.00 & \textbf{68.33} & \textbf{51.66} & 115.00 & 409.99 \\
 & IMCCD & \textbf{185.00} & 55.00 &50.00  &100.00& 390.00 \\
 & CLVA (ours) & \textbf{185.00} & 55.00 & 50.00 &\textbf{130.00}  & \textbf{420.00} \\
 \midrule
 \multirow{6}{*}{Qwen-VL} & Greedy & 175.00 & 140.00 & 123.33 & 180.00 & 618.33 \\
 & Beam Search &180.00  & 141.66 &  133.33& 180.00 & 634.99 \\
 & MemVR & 185.00 &145.00  & 123.33 & \textbf{185.00} & 638.33 \\
 & ClearSight &185.00  & 141.66 & 148.33 & 180.00 & 654.99 \\
 & IMCCD & \textbf{190.00} & 140.00 & 138.33 & 175.00 & 643.33 \\
 & CLVA (ours) &180.00  & \textbf{146.66} & \textbf{158.33} & \textbf{185.00} & \textbf{669.99} \\
\bottomrule
\end{tabular}
}
\end{table}

The results in Table \ref{tab:pope} show that CLVA achieves a performance boost compared to other state of the art training free methods. Our approach yields significant improvements in both Accuracy(ranging from 0.59\% to 7.13\%) and F1 score(ranging from 0.28\% to 6.33\%). Beyond standard architectures, CLVA also achieves substantial performance enhancements when applied to models utilizing Q former architectures. This demonstrates that our intervention mechanism is not limited to a specific framework but possesses broad applicability across different model designs. While some results for CLVA do not surpass those of IMCCD on the POPE benchmark, our method offers a decisive advantage in terms of computational efficiency. CLVA requires only half the video memory compared to IMCCD and operates at nearly twice the speed during inference. For further details, please refer to Appendix \ref{A}.

\textbf{CHAIR.} This benchmark provides a systematic framework for evaluating object hallucination in image captioning tasks. Caption accuracy is assessed by comparing mentioned objects against ground truth labels. A hallucination is defined as an object present in the generated description but absent from the ground truth. Following standard experimental protocols, we evaluate on 500 randomly sampled images from the MSCOCO 2014 validation set. The model is prompted to describe the image in detail with a maximum generation length set to 128 tokens. Lower values in these metrics indicate a higher degree of visual grounding and fewer hallucinations.


\begin{figure*}[t]
\centering
\includegraphics[width=\linewidth]{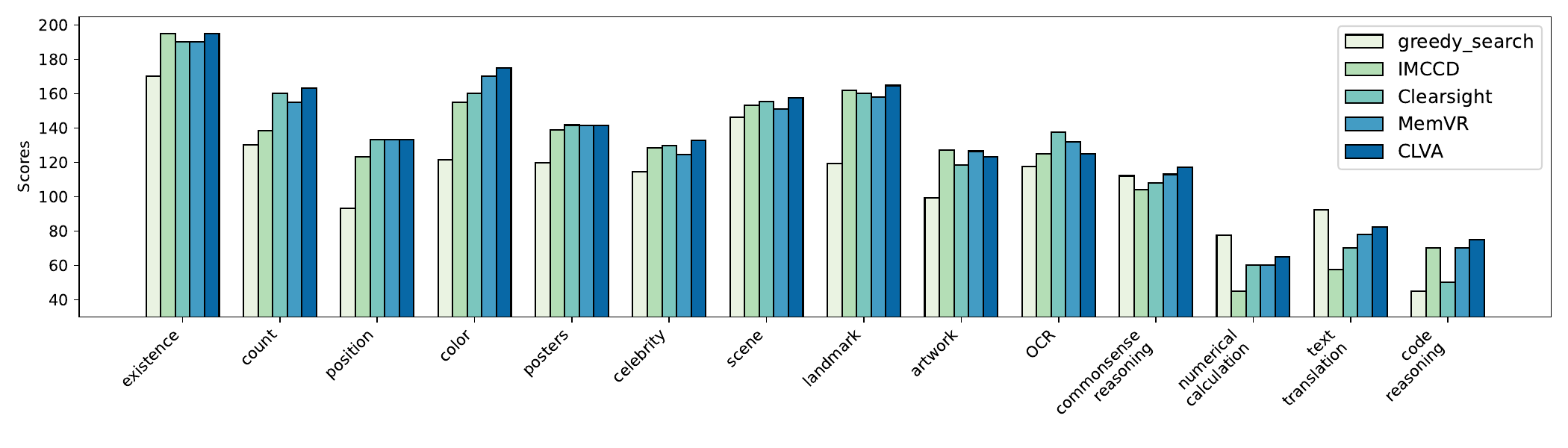} 
\caption{Results of LLaVA1.5 on MME-Fullset.}
\label{fig:mmefull}
\end{figure*}

The experimental results in Table \ref{tab:chair} demonstrate that CLVA yields a significant reduction in hallucination rates across different model architectures. For the LLaVA model, $CHAIR_s$ and $CHAIR_i$ are reduced by 5\% and 2.5\%, respectively. Similarly, InstructBLIP exhibits even more substantial gains with reductions of 6.8\% in $CHAIR_s$ and 2.7\% in $CHAIR_i$. Beyond these primary metrics, all models also show improved performance in Recall, indicating that CLVA successfully reduces hallucinations without sacrificing the descriptive capacity of the models. This performance is particularly evident in long descriptions where attention drift in deep layers typically causes the model to lose focus on visual features. These findings confirm the ability of the method to achieve superior visual grounding and consistency during the auto regressive generation process. The scores achieved by our method highlight the capacity of CLVA to produce descriptions that are both detailed and factually accurate.
\subsection{General Benchmark Results}
We further evaluate CLVA on general benchmarks to ensure that the suppression of hallucinations does not lead to performance degradation in other core capabilities. We select MME\cite{fu2025mme} as the primary benchmark to assess the balance between perception reliability and general visual understanding.

\textbf{MME.} We report the performance of CLVA on the MME benchmark, which provides a comprehensive assessment of both perception and cognition abilities. To offer a detailed view of the model reliability, we include the total scores for the entire MME collection alongside the individual results for the four specific hallucination subsets. These subsets(existence, count, position, and color) are designed to probe the tendency of the model to generate non existent or incorrect visual details.

We present the results on the MME benchmark in Table \ref{tab:mme}. We find that all three models show an increase in total scores across the MME hallucination subsets. We observe a particularly significant improvement for LLaVA 1.5, where the score on these subsets rises from 514 to 666. We also evaluate the performance across all 14 subtasks as illustrated in Figure \ref{fig:mmefull}. We note that the total MME scores exhibit an upward trend after applying CLVA. Specifically, LLaVA-1.5 shows consistent improvements across all perception capabilities. The performance in numerical calculation and text translation decreases because these tasks primarily involve language decoding rather than visual grounding. These results demonstrate that our method effectively mitigates hallucination phenomena and enhances the overall performance of the models.

\subsection{Ablation Studies}

Ablation experiments are conducted on the POPE MS-COCO subset and the MME dataset to evaluate the individual contributions of POS and NEG anchors within CLVA. As shown in Table \ref{tab:ablation}, the results demonstrate that both POS and NEG anchors are essential for hallucination mitigation.

The configuration utilizing vision sensitive heads as positive anchors provides the necessary visual grounding to pull the attention of deep layers back to the correct visual regions. Simultaneously, using insensitive heads as negative anchors successfully suppresses the regressive noise originating from initial layers. The final row of Table \ref{tab:ablation} represents the negative addition of anchors, where negative anchors from insensitive heads are strengthened while positive anchors from vision sensitive heads are weakened. The significant performance degradation in this setting confirms that the specific orientation of our cross layer mechanism is critical for maintaining visual fidelity. In addition, we conduct further ablation studies on different parameters, with detailed results and analyses provided in the Appendix \ref{A}.
\begin{table}[t]
\caption{Ablation study of different components in CLVA for LLaVA-1.5. POS and NEG denote positive and negative anchors, while + and - represent the positive and negative usage of anchors, respectively.}
\label{tab:ablation}
\resizebox{\columnwidth}{!}{
\begin{tabular}{cccccc}
\toprule
\multirow{2}{*}{\textbf{POS}}  & \multirow{2}{*}{\textbf{NEG}}  & \textbf{MME} & \textbf{MME} & \multicolumn{2}{c}{\textbf{POPE(Random \& COCO)}} \\
& & \textbf{(Hallucination)} & \textbf{(Overall)} & \textbf{F1-score} & \textbf{Accuracy}\\
\midrule
 & &514.99 &1558.76 &84.60 &86.76\\
\Large\textbf{+} & &654.66 &1843.92 &85.39 &87.00\\
& \Large\textbf{+}&643.66 & 1819.95&85.50 &87.06\\
\Large\textbf{+}  & \Large\textbf{+}& 666.66& 1851.23& 86.99&88.26\\
\Large\textbf{-}  & \Large\textbf{-}&505.00 & 1531.26& 83.73 & 86.20\\

\bottomrule
\end{tabular}
}
\end{table}

\section{Conclusion}
This study investigates the layer wise evolution of visual features and discovers that hallucination in multimodal models is closely linked to deep layer attention drift and the regression toward initial noise from insensitive heads. We propose CLVA, the Cross Layer Visual Anchor method, which is a training-free solution that reinforces critical mid-layer features and pulls drifting attention back to the correct visual regions. Our results across diverse architectures and benchmarks demonstrate that CLVA effectively mitigates hallucination and enhances perception performance without adding significant computational or memory overhead. This work provides a robust and efficient mechanism for improving the visual grounding of vision language systems.


\section*{Impact Statement}

This paper presents work whose goal is to advance the field of Machine
Learning. There are many potential societal consequences of our work, none
which we feel must be specifically highlighted here.

\bibliography{icml2026}
\bibliographystyle{icml2026}
\newpage
\appendix
\onecolumn
\section{Additional Experimental Results}
\label{A}
\subsection{Speed and Memory Comparison}
\label{A1}
We conduct our experiments on a single NVIDIA H20 GPU. We compare the resource consumption of our proposed CLVA and IMCCD on the POPE dataset, as shown in Table \ref{tab:Memory}. We find that CLVA introduces only minimal additional overhead compared to the original baseline model, which demonstrates the clear advantage of our method in terms of resource consumption. Therefore, although some of our experimental results do not exceed those of IMCCD, our advantages in terms of inference speed and memory efficiency remain very large. When the difference is the greatest, the memory we require is only half of that needed for IMCCD, while our speed is about twice as fast.

\begin{table}[ht]
\centering
\caption{Comparison of inference speed and GPU memory usage for different models and methods on the POPE dataset}
\label{tab:Memory}
\begin{tabular}{llcccc}
\toprule
Model & Method & Accuracy & Total Time & GPU-Memory & Latency/Example \\ \midrule
\multirow{3}{*}{LLaVA-1.5} & Greedy & 84.79 & 0:7:48 & 15.7G & 0.155s \\
 & IMCCD & 85.49 & 0:16:38 & 18.7G & 0.353s \\
 \rowcolor[gray]{0.9}  & CLVA (ours) & 85.98 & 0:7:54 & 15.8G & 0.157s \\ 
 \midrule
\multirow{3}{*}{InstructBLIP} & Greedy & 77.02 & 0:5:07 & 16.6G & 0.102s \\
 & IMCCD & 81.60 & 0:10:30 & 16.7G & 0.207s \\
\rowcolor[gray]{0.9}  & CLVA (ours) & 82.46 & 0:5:22 & 16.6G & 0.105s \\ 
\midrule
\multirow{3}{*}{Qwen-VL} & Greedy & 84.62 & 0:14:34 & 20.4G & 0.296s \\
 & IMCCD & 86.56 & 0:35:22 & 24.9G & 0.709s \\
\rowcolor[gray]{0.9}  & CLVA (ours) & 86.77  & 0:15:45 & 20.4G & 0.321s \\ 
\bottomrule
\end{tabular}
\end{table}

\subsection{Ablation Study}
\label{A2}
\begin{figure*}[h]
\centering
\includegraphics[width=\linewidth]{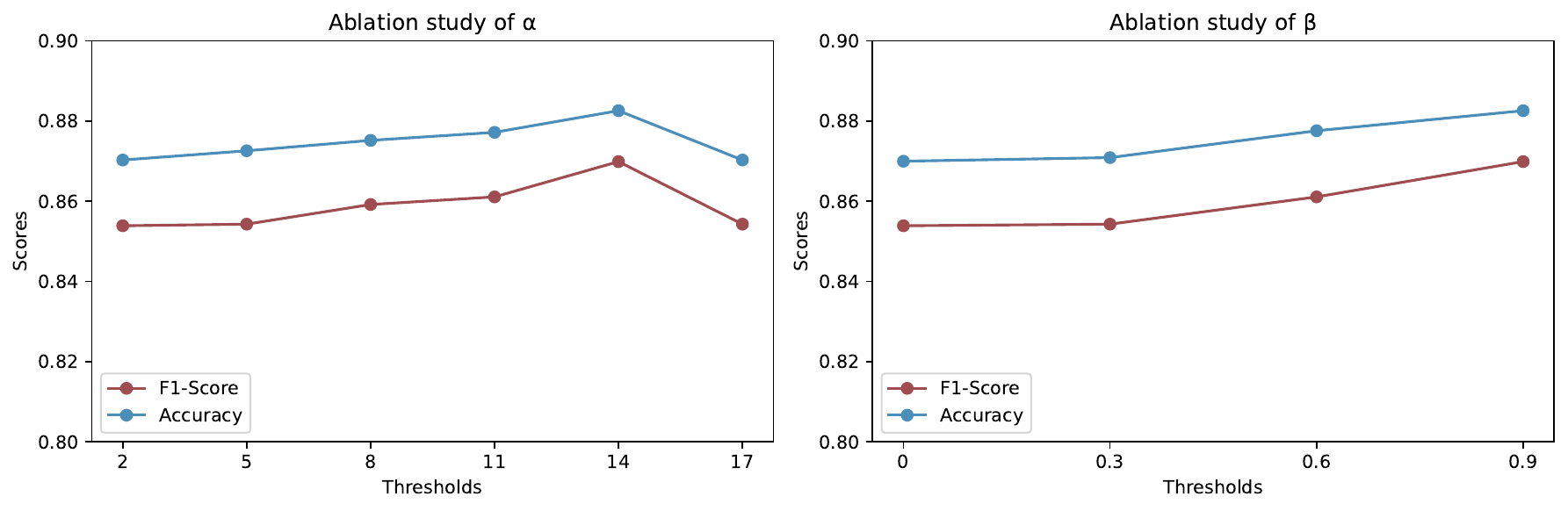} 
\caption{Ablation study of $\alpha$ and $\beta$}
\label{fig:a}
\end{figure*}

We conduct parameter sensitivity experiments for $\alpha$ and $\beta$ on the POPE random coco subset, as shown in Figure \ref{fig:a}. We observe that when the positive anchor modification parameter $\alpha$ exceeds 14, the model performance begins to decline. Similarly, the performance of the model decreases as the negative anchor correction parameter $\beta$ provides less suppression. We believe the performance drop associated with an excessively large $\alpha$ occurs because the model focuses too intensely on visual regions, which eventually disrupts the original understanding of the overall semantics. This demonstrates that while anchoring is effective, maintaining a balance between visual focus and linguistic context is crucial for maintaining the robustness of the autoregressive generation process.

\subsection{Experimental Results with Nucleus Sampling}
We document the performance of CLVA during nucleus sampling on the coco subset of POPE. As shown in Table \ref{tab:coco}, we include these findings in the appendix to provide a more comprehensive view, as stochastic decoding typically lacks the high reproducibility associated with greedy search. Our evaluation spans three separate models to ensure the broad applicability of our findings. We observe that CLVA consistently maintains a clear performance margin over all baseline methods under these sampling conditions. This sustained advantage is a direct result of the double anchors mechanism.

\begin{table}[ht]
\centering
\caption{Results of different models using nucleus sampling on the POPE dataset}
\label{tab:coco}
\begin{tabular}{llcccccc}
\toprule
 & \multirow{2}{*}{Method}& \multicolumn{2}{c}{LLaVA-1.5} & \multicolumn{2}{c}{InstructBLIP} & \multicolumn{2}{c}{Qwen-VL} \\ \cmidrule(lr){3-4} \cmidrule(lr){5-6} \cmidrule(lr){7-8}
 &  & F1 & Accuracy & F1 & Accuracy & F1 & Accuracy \\ \midrule
\multirow{2}{*}{random} 
& sample & 81.94 & 83.76 & 81.12 & 81.50 & 83.72 & 85.73 \\
& ours & 82.97 & 84.33 & 84.06 & 84.43 & 84.87 & 86.56 \\
\multirow{2}{*}{popular} 
& sample & 80.86 & 82.56 & 78.74 & 78.50 & 83.03 & 84.96 \\
& ours & 83.52 & 84.96 & 82.54 & 81.55 & 83.54 & 85.86 \\
\multirow{2}{*}{adversarial} 
& sample & 78.43 & 79.73 & 77.91 & 77.43 & 81.43 & 83.13 \\
& ours & 79.83 & 80.76 & 80.78 & 80.30 & 83.04 & 84.53 \\ \bottomrule
\end{tabular}
\end{table}

\subsection{Visualization of Attention Heatmaps}
\begin{figure*}[h]
\centering
\includegraphics[width=\linewidth]{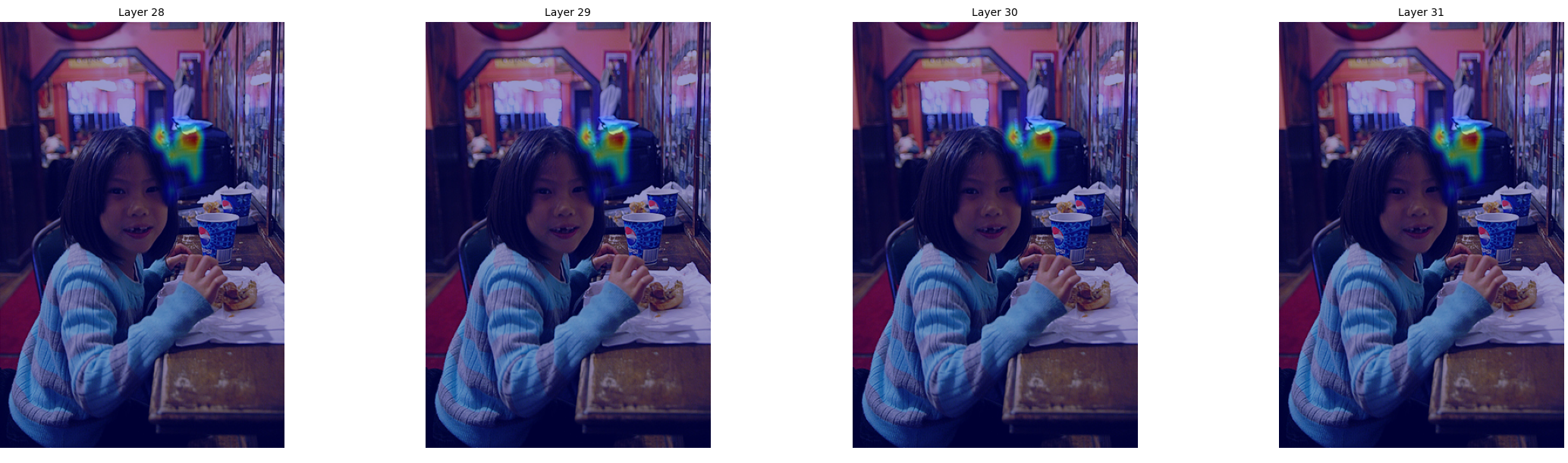} 
\caption{Is there a backpack in the image?}
\label{fig:backpack}
\end{figure*}

\begin{figure*}[h]
\centering
\includegraphics[width=\linewidth]{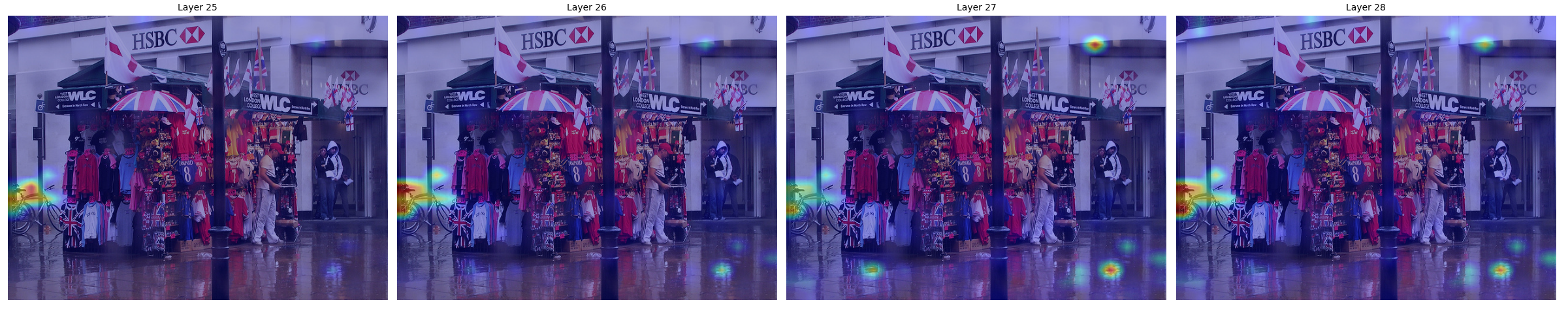} 
\caption{Is there a bicycle in the image?}
\label{fig:bicycle}
\end{figure*}

We provide visualizations of the attention heatmaps from the subsequent layers of the CLVA-weighted model to demonstrate the effectiveness of our weighting mechanism. We observe that the significant attention drift typically found in deeper layers is successfully corrected through the application of CLVA. As illustrated in Figure \ref{fig:bicycle} and Figure \ref{fig:backpack}, our method pulls the drifted attention back to the precise visual regions that were previously ignored or misaligned. We achieve this spatial recalibration through the double anchors mechanism, which redirects the focus of the model toward relevant image features. By leveraging the visual guidance and suppressing the noise from insensitive heads, we ensure that the model maintains a strong and accurate visual grounding during the autoregressive generation process.

\subsection{Case study on LLaVA-Bench}
\begin{figure*}[h]
\centering
\includegraphics[width=\linewidth]{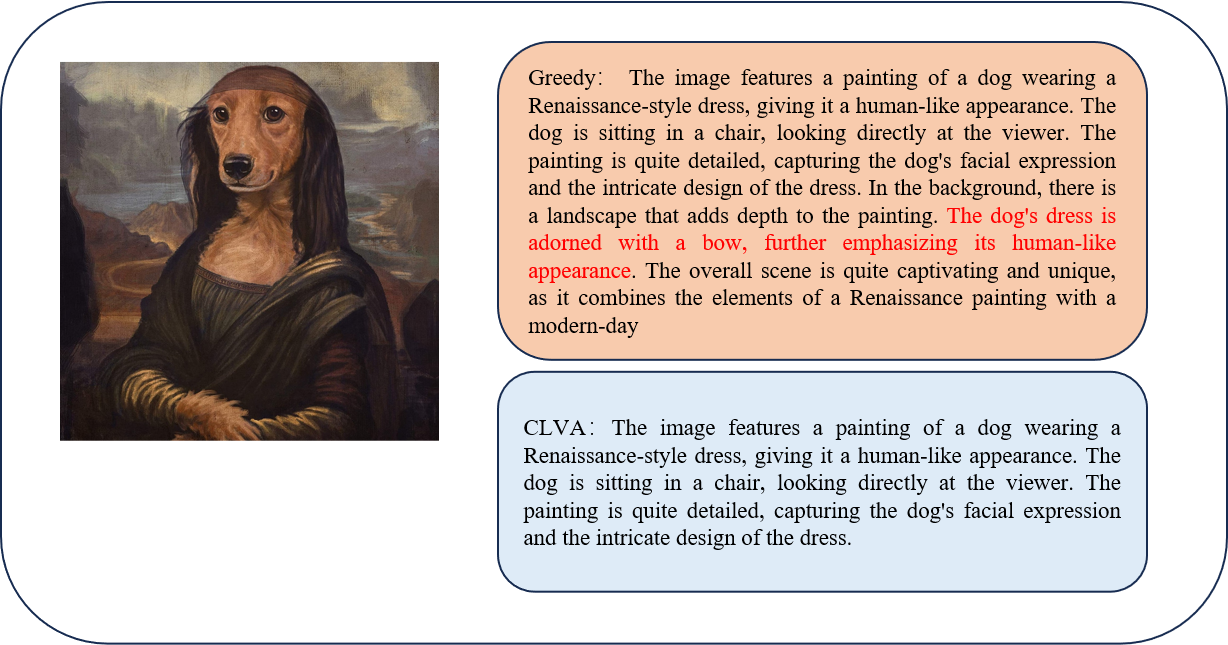} 
\caption{An example of the LLaVA-1.5 model on LLaVA-Bench; red text indicates incorrect answers.}
\label{fig:003}
\end{figure*}

\begin{figure*}[h]
\centering
\includegraphics[width=\linewidth]{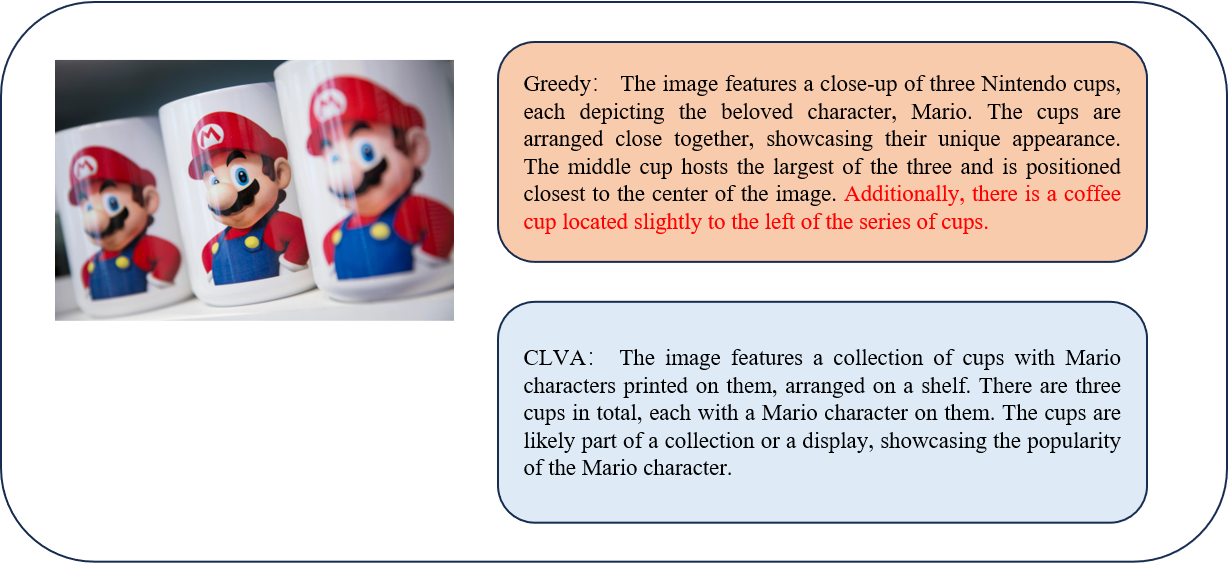} 
\caption{An example of the InstructBLIP model on LLaVA-Bench; red text indicates incorrect answers.}
\label{fig:012}
\end{figure*}

\begin{figure*}[h]
\centering
\includegraphics[width=\linewidth]{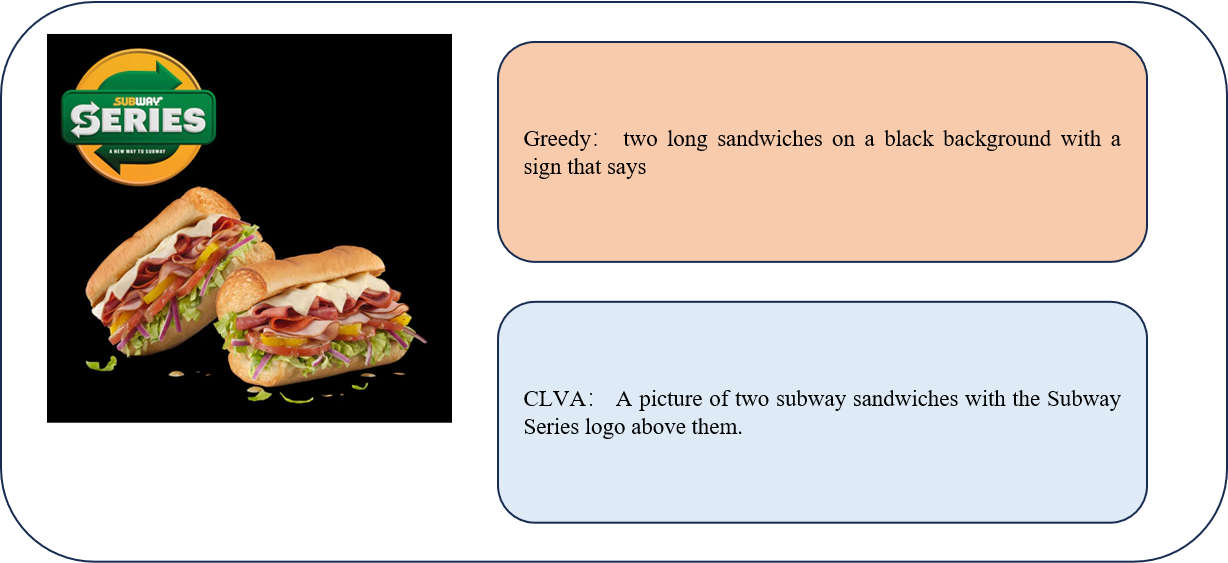} 
\caption{An example of the Qwen-VL model on LLaVA-Bench; red text indicates incorrect answers.}
\label{fig:023}
\end{figure*}

We evaluate CLVA on the LLaVA-Bench dataset using the LLaVA-1.5 model to assess its performance in more complex and open ended scenarios. As shown in (Figure \ref{fig:003}, Figure \ref{fig:012}, Figure \ref{fig:023}). LLaVA Bench is designed to test the capabilities of models in real world settings, focusing on detailed descriptions, complex reasoning, and conversational logic. We compare our results with the original baseline model to demonstrate the practical benefits of the double anchors mechanism. We find that CLVA consistently produces more accurate and grounded responses than the baseline. This improvement is particularly evident in long generation tasks where the model must maintain a consistent focus on specific image regions. By utilizing insensitive heads to mitigate the impact of linguistic priors, we show that our training-free method effectively prevents the model from generating hallucinations during the autoregressive process, leading to a higher degree of alignment between the visual input and the generated text.

\section{Theoretical Analysis}
\subsection{Analysis of Q-former Architectures}
We investigate the theoretical limitations of applying attention modifications directly to the compressed output of a Qformer. In architectures like BLIP-2, the Q-former compresses a large set of raw visual features into a significantly smaller set of latent tokens $Z$. When the LLM performs cross-attention over these tokens, the spatial information is already highly diffused. We find that applying the double anchors mechanism at the LLM decoder level for these compressed tokens leads to suboptimal results. The core of our modification follows the weighting formula:
\begin{equation}
\tilde{\mathbf{A}}(i, j) = \mathbf{A}(i, j) \cdot \left(1+\alpha \mathbf{Z}_{pos}(j) - \beta \mathbf{Z}_{neg}(j) \right)
\end{equation}
where $\mathbf{A}(i, j)$ represents the original attention weight, and $\mathbf{Z}_{pos}$ and $\mathbf{Z}_{neg}$ are the positive visual anchors and negative noise anchors derived from insensitive heads. The primary theoretical challenge is that each compressed token $z_j$ no longer maintains a one to one correspondence with a specific spatial region $(x, y)$ in the original image. Instead, each token is a high-level semantic aggregator. If we were to apply the above formula to the LLM attention weights $\mathbf{A}(i, j)$ targeting these compressed tokens, the amplification or suppression would act on a "blurred" semantic representation. This leads to a loss of spatial precision and disrupts the delicate balance of the latent space, which we can characterize as a reduction in the mutual information between the original image and the weighted tokens. To overcome this, we adapt the double anchors mechanism by applying the weighting formula directly within the internal cross-attention layers of the Qformer itself. During the compression phase, the Qformer queries $Q$ interact with the raw image features $K, V$. By applying the modification:
\begin{equation}
\tilde{\mathbf{A}}_{Qf}(i, j) = \mathbf{A}_{Qf}(i, j) \cdot \left(1+\alpha \mathbf{Z}_{pos}(j) - \beta \mathbf{Z}_{neg}(j) \right)
\end{equation}
We ensure that the latent queries are forced to aggregate information from the relevant visual regions while actively suppressing background or linguistic noise before the compression is finalized. This internal application of double anchors ensures that the resulting compressed tokens provided to the LLM are spatially "cleaner" and semantically richer. Our experiments confirm that this architectural adaptation leads to a significant performance boost, as it mitigates the attention drift at the source of the visual representation rather than attempting to correct it after the spatial information has already been compressed.

\subsection{Motivation for Distinguishing Attention Heads}

We analyze the necessity of separating attention heads into functional categories rather than applying a global average for anchor extraction. Our theoretical framework relies on the observation that attention heads in Large Multi Modal Models exhibit high degrees of functional specialization. Specifically, vision sensitive heads focus on grounded object features, while insensitive heads primarily process linguistic transitions or global background noise. We argue that a simple averaging of all heads $\mathbf{Z}_{avg} = \frac{1}{H} \sum_{h=1}^H \mathbf{Z}_h$ acts as a low pass filter that blurs these distinct signals. If we were to replace our specific anchors with a global average in the weighting formula:
\begin{equation}
\tilde{\mathbf{A}}(i, j) = \mathbf{A}(i, j) \cdot \left(1+(\alpha - \beta) \mathbf{Z}_{avg}(j) \right)
\end{equation}
the contrastive power of the mechanism is lost. In this scenario, the positive signal from sensitive heads is diluted by the noise from insensitive heads, while the negative anchor $\mathbf{Z}_{neg}$ becomes "contaminated" with actual visual information. This leads to a significant reduction in the signal to noise ratio of the visual guidance.

By isolating insensitive heads, we obtain a "pure" baseline that represents the internal linguistic bias of the model. This allows the negative anchor $\mathbf{Z}_{neg}$ to precisely identify which regions are being attended to due to auto-regressive language patterns rather than actual visual evidence. Simultaneously, extracting $\mathbf{Z}_{pos}$ from sensitive heads ensures that the positive reinforcement is concentrated on high value visual features.

We conclude that this distinction is critical for the success of the double anchors mechanism. It creates a contrastive effect that allows our weighting formula to effectively "push" the attention of the model away from linguistic hallucinations and "pull" it toward grounded visual tokens. Without this separation, the model would fail to distinguish between meaningful visual cues and the inherent noise of the deep layers, ultimately leading to the very attention drift that CLVA is designed to prevent.

\end{document}